\documentclass[11pt,letterpaper]{article}
\usepackage{authblk}
\usepackage[pdfborder={0 0 0}]{hyperref}
\usepackage{emnlp2016}
\usepackage{times}
\usepackage{latexsym}
\usepackage{booktabs}
\usepackage{xcolor}
\usepackage{cleveref}
\usepackage{microtype}
\usepackage[utf8]{inputenc}
\crefname{section}{§}{§§}
\Crefname{section}{§}{§§}

\emnlpfinalcopy 
\def\emnlppaperid{***} 

\definecolor{limegreen}{rgb}{0.2, 0.8, 0.2}

\usepackage[english]{babel}
\usepackage[T1]{fontenc}
\usepackage{graphicx}
\usepackage[colorinlistoftodos]{todonotes}
\usepackage{todonotes}

\title{SimVerb-3500: A Large-Scale Evaluation Set of Verb Similarity}

\author[1]{Daniela Gerz}
\author[1]{Ivan Vuli\'{c}}
\author[1]{Felix Hill}
\author[2]{Roi Reichart}
\author[1]{Anna Korhonen}
\affil[1]{Language Technology Lab, DTAL, University of Cambridge}
\affil[2]{Faculty of Industrial Engineering and Management, Technion, IIT}
\date{\today}

\begin{document}
\maketitle
%
\begin{abstract}

Verbs play a critical role in the meaning of sentences, but these ubiquitous words have received little attention in recent distributional semantics research. 
We introduce SimVerb-3500, an evaluation resource that provides human ratings for the similarity of 3,500 verb pairs.  SimVerb-3500 covers all normed verb types from the USF free-association database, providing at least three examples for every VerbNet class. This broad coverage facilitates detailed analyses of how syntactic and semantic phenomena together influence human understanding of verb meaning. Further, with significantly larger development and test sets than existing benchmarks, SimVerb-3500 enables more robust evaluation of representation learning architectures and promotes the development of methods tailored to verbs. We hope that SimVerb-3500 will enable a richer understanding of the diversity and complexity of verb semantics and guide the development of systems that can effectively represent and interpret this meaning.

\end{abstract}

\section{Introduction}
\label{sec:intro}

Verbs are famously both complex and variable. They express the semantics of an event as well the relational information among participants in that event, and they display a rich range of syntactic and semantic behaviour \cite{Jackendoff:1972book,Gruber:1976book,Levin:1993book}. Verbs play a key role at almost every level of linguistic analysis. Information related to their predicate argument structure can benefit many NLP tasks (e.g. parsing, semantic role labelling, information extraction) and applications (e.g. machine translation, text mining) as well as research on human language acquisition and processing \cite{Korhonen2010royalsociety}. Precise methods for representing and understanding verb semantics will undoubtedly be necessary for machines to interpret the meaning of sentences with similar accuracy to humans. 

Numerous algorithms for acquiring word representations from text and/or more structured knowledge bases have been developed in recent years \cite{Mikolov:2013iclr,Pennington:2014emnlp,Faruqui:2015naacl}. These representations (or \emph{embeddings}) typically contain powerful features that are applicable to many language applications \cite{Collobert:2008icml,Turian:2010acl}. Nevertheless, the predominant approaches to distributed representation learning apply a single learning algorithm and representational form for all words in a vocabulary. This is despite evidence that applying different learning algorithms to word types such as nouns, adjectives and verbs can significantly increase the ultimate usefulness of representations \cite{Schwartz:2015conll}. 

One factor behind the lack of more nuanced word representation learning methods is the scarcity of satisfactory ways to evaluate or analyse representations of particular word types. Resources such as MEN \cite{Bruni:2014jair}, Rare Words \cite{Luong:2013conll} and SimLex-999 \cite{Hill:2015cl} focus either on words from a single class or small samples of different word types, with automatic approaches already reaching or surpassing the inter-annotator agreement ceiling. Consequently, for word classes such as \emph{verbs}, whose semantics is critical for language understanding, it is practically impossible to achieve statistically robust analyses and comparisons between different representation learning architectures.

To overcome this barrier to verb semantics research, we introduce {\em SimVerb-3500} -- an extensive intrinsic evaluation resource that is unprecedented in both size and coverage. SimVerb-3500 includes 827 verb types from the University of South Florida Free Association Norms (USF) \cite{Nelson:2004usf}, and at least 3 member verbs from each of the 101 top-level VerbNet classes \cite{Kipper:2008lre}. This coverage enables researchers to better understand the complex diversity of syntactic-semantic verb behaviours, and provides direct links to other established semantic resources such as WordNet \cite{Miller:1995cacm} and PropBank \cite{Palmer:2005cl}. Moreover, the large standardised development and test sets in SimVerb-3500 allow for principled tuning of hyperparameters, a critical aspect of achieving strong performance with the latest representation learning architectures.

In \cref{s:rw}, we discuss previous evaluation resources targeting verb similarity. We present the new SimVerb-3500 data set along with our design choices and the pair selection process in \cref{s:dataset}, while the annotation process is detailed in \cref{s:annotation}. In \cref{s:analysis} we report the performance of a diverse range of popular representation learning architectures, together with benchmark performance on existing evaluation sets.  In \cref{s:evaluation}, we show how SimVerb-3500 enables a variety of new linguistic analyses, which were previously impossible due to the lack of coverage and scale in existing resources.

\section{Related Work}
\label{s:rw}
A natural way to evaluate representation quality is by judging the similarity of representations assigned to similar words. The most popular evaluation sets at present consist of word pairs with similarity ratings produced by human annotators.\footnote{In some existing evaluation sets pairs are scored for relatedness which has some overlap with similarity. SimVerb-3500 focuses on similarity as this is a more focused semantic relation that seems to yield a higher agreement between human annotators. For a broader discussion see \cite{Hill:2015cl}.}
Nevertheless, we find that all available datasets of this kind are insufficient for judging verb similarity 
due to their small size or narrow coverage of verbs. 

In particular, a number of word pair evaluation sets are prominent in the distributional semantics literature. 

Representative examples include RG-65 \cite{Rubenstein1965acm} and WordSim-353 \cite{Finkelstein:2002tois,Agirre:2009naacl} which are small (65 and 353 word pairs, respectively). Larger evaluation sets such as the Rare Words evaluation set \cite{Luong:2013conll} (2034 word pairs) and the evaluations sets from \newcite{Silberer:2014acl} are dominated by noun pairs and the former also focuses on low-frequency phenomena. Therefore, these datasets do not provide a representative sample of verbs \cite{Hill:2015cl}.

Two datasets that do focus on verb pairs to some extent are the data set of \newcite{Baker:2014emnlp} and Simlex-999 \cite{Hill:2015cl}. These datasets, however, still contain a limited number of verb pairs (134 and 222, respectively), making them unrepresentative of the rich variety of verb semantic phenomena.

In this paper we provide a remedy for this problem by presenting a more comprehensive and representative verb pair evaluation resource. 

\section{The SimVerb-3500 Data Set}
\label{s:dataset}

In this section, we discuss the design principles behind SimVerb-3500.  
We first demonstrate that a new evaluation resource for verb similarity is a necessity. We then describe how the final verb pairs were selected with the goal to be representative, that is, to guarantee a wide coverage of two standard semantic resources: USF and VerbNet.

\subsection{Design Motivation}
\newcite{Hill:2015cl} argue that comprehensive high-quality evaluation resources have to satisfy the following three criteria: {\em (C1) Representative} (the resource covers the full range of concepts occurring in natural language); {\em (C2) Clearly defined} (it clearly defines the annotated relation, e.g., similarity); (C3) {\em Consistent and reliable} (untrained native speakers must be able to quantify the target relation consistently relying on simple instructions). 

Building on the same annotation guidelines as Simlex-999 that explicitly targets similarity, we ensure that criteria C2 and C3 are satisfied. However, even SimLex, as the most extensive evaluation resource for verb similarity available at present, is still of limited size, spanning only 222 verb pairs and 170 distinct verb lemmas in total. Given that 39 out of the 101 top-level VerbNet classes are not represented at all in SimLex, while 20 classes have only one member verb,\footnote{Note that verbs in VerbNet are soft clustered, and one verb type may be associated with more than one class. When computing coverage, we assume that such verbs attribute to counts of all their associated classes.} one may conclude that the criterion C1 is not at all satisfied with current resources.

There is another fundamental limitation of all current verb similarity evaluation resources: automatic approaches have reached or surpassed the inter-annotator agreement ceiling. For instance, while the average pairwise correlation between annotators on SL-222 is Spearman's $\rho$ correlation of $0.717$, the best performing automatic system reaches $\rho=0.727$ \cite{Mrksic:2016naacl}. SimVerb-3500 does not inherit this anomaly (see Tab.~\ref{tab:gaps}) and demonstrates that there still exists an evident gap between the human and system performance.

In order to satisfy C1-C3, the new SimVerb-3500 evaluation set contains similarity ratings for 3,500 \textit{verb pairs}, containing 827 verb types in total and 3 member verbs for each top-level VerbNet class. The rating scale goes from 0 (not similar at all) to 10 (synonymous). We employed the SimLex-999 annotation guidelines. In particular, we instructed annotators to give low ratings to antonyms, and to distinguish between similarity and relatedness. Pairs that are related but not similar (e.g., \textit{to snore / to snooze}, \textit{to walk / to crawl}) thus have a fairly low rating. Several example pairs are provided in Tab.~\ref{tab:examples}.

\subsection{Choice of Verb Pairs and Coverage}
To ensure a wide coverage of a variety of syntactico-semantic phenomena (C1), the choice of verb pairs is steered by two standard semantic resources available online: (1) the USF norms data set\footnote{http://w3.usf.edu/FreeAssociation/} \cite{Nelson:2004usf}, and (2) the VerbNet verb lexicon\footnote{http://verbs.colorado.edu/verb-index/} \cite{Kipper:2004lrec,Kipper:2008lre}.

The {\bf USF} norms data set (further USF) is the largest database of free association collected for English. It was generated by presenting human subjects with one of $5,000$ cue concepts and asking them to write the first word coming to mind that is associated with that concept. Each cue concept $c$ was normed in this way by over 10 participants, resulting in a set of associates $a$ for each cue, for a total of over $72,000$ $(c,a)$ pairs. For each such pair, the proportion of participants who produced associate $a$ when presented with cue $c$ can be used as a proxy for the strength of association between the two words. 

The norming process guarantees that two words in a pair have a degree of semantic association which correlates well with semantic relatedness and similarity. Sampling from the USF set ensures that both related but non-similar pairs (e.g., {\em to run / to sweat}) as well as similar pairs (e.g., {\em to reply / to respond}) are represented in the final list of pairs. Further, the rich annotations of the output USF data (e.g., concreteness scores, association strength) can be directly combined with the SimVerb-3500 similarity scores to yield additional  analyses and insight.

\begin{table}[!t]
\begin{center}
\def\arraystretch{0.93}
\begin{small}
\begin{tabular}{ l | r }
  Pair & Rating \\ \hline
  to reply / to respond & 9.79 \\
  to snooze / to nap & 8.80 \\
  to cook / to bake & 7.80 \\
  to participate / to join & 5.64 \\
  to snore / to snooze & 4.15 \\
  to walk / to crawl & 2.32 \\
  to stay / to leave & 0.17 \\
  to snooze / to happen & 0.00 \\
\end{tabular}
\end{small}
\end{center}
\vspace{-0em}
\caption{Example verb pairs from SimVerb-3500.}
\vspace{-1.2em}
\label{tab:examples}
\end{table}

{\bf VerbNet} ({\bf VN}) is the largest online verb lexicon currently available for English. It is hierarchical, domain-independent, and broad-coverage. VN is organised into verb classes extending the classes from \newcite{Levin:1993book} through further refinement to achieve syntactic and semantic coherence among class members. According to the official VerbNet guidelines,\footnote{{\scriptsize http://verbs.colorado.edu/verb-index/VerbNet\_Guidelines.pdf}} ``Verb Classes are numbered according to shared semantics and syntax, and classes which share a top-level number	(9-109)	have corresponding semantic relationships.''  For instance, all verbs from the top-level Class 9 are labelled ``Verbs of Putting'', all verbs from Class 30 are labelled ``Verbs of Perception'', while Class 39 contains ``Verbs of Ingesting''. 

Among others, three basic types of information are covered in VN: (1) verb subcategorization frames (SCFs), which describe the syntactic realization of the predicate-argument structure (e.g. {\em The window broke}), (2) selectional preferences (SPs), which capture the semantic preferences verbs have for their arguments (e.g. {\em a breakable physical object} broke) and (3) lexical-semantic verb classes (VCs) which provide a shared level of abstraction for verbs similar in their (morpho-)syntactic and semantic properties (e.g. {\em BREAK verbs}, sharing the VN class 45.1, and the top-level VN class 45).\footnote{https://verbs.colorado.edu/verb-index/vn/break-45.1.php} The basic overview of the VerbNet structure already suggests that measuring verb similarity is far from trivial as it revolves around a complex interplay between various semantic and syntactic properties.

The wide coverage of VN in SimVerb-3500 assures the wide coverage of distinct verb groups/classes and their related linguistic phenomena. Finally, VerbNet enables further connections of SimVerb-3500 to other important lexical resources such as FrameNet \cite{Baker:1998acl}, WordNet \cite{Miller:1995cacm}, and PropBank \cite{Palmer:2005cl} through the sets of mappings created by the SemLink project initiative \cite{Loper:2007ws}.\footnote{https://verbs.colorado.edu/semlink/}

\paragraph{Sampling Procedure}
We next sketch the complete sampling procedure which resulted in the final set of 3500 distinct verb pairs finally annotated in a crowdsourcing study (\cref{s:annotation}). 

\noindent {\bf (Step 1)} We extracted all possible verb pairs from USF based on the associated POS tags available as part of USF annotations. To ensure that semantic association between verbs in a pair is not accidental, we then discarded all such USF pairs that had been associated by 2 or less participants in USF.

\noindent {\bf (Step 2)} We then manually cleaned and simplified the list of pairs by removing all pairs with multi-word verbs (e.g., {\em quit / give up}), all pairs that contained the non-infinitive form of a verb (e.g., {\em accomplished / finished}, {\em hidden / find}), removing all pairs containing at least one auxiliary verb (e.g., {\em must / to see, must / to be}). The first two steps resulted in 3,072 USF-based verb pairs. 

\noindent {\bf (Step 3)} After this stage, we noticed that several top-level VN classes are not part of the extracted set. For instance, 5 VN classes did not have any member verbs included, 22 VN classes had only 1 verb, and 6 VN classes had 2 verbs included in the current set.

We resolved the VerbNet coverage issue by sampling from such 'under-represented' VN classes directly. Note that this step is not related to USF at all. For each such class we sampled additional verb types until the class was represented by 3 or 4 member verbs (chosen randomly).\footnote{The following three VN classes are exceptions: (1) Class 56, consisting of words that are dominantly tagged as nouns, but can be used as verbs exceptionally (e.g., {\em holiday, summer, honeymoon}); (2) Class 91, consisting of 2 verbs ({\em count, matter}); (3) Class 93, consisting of 2 single word verbs ({\em adopt, assume}).} Following that, we sampled at least 2 verb pairs for each previously 'under-represented' VN class by pairing 2 member verbs from each such class. This procedure resulted in 81 additional pairs, now 3,153 in total. 

\noindent {\bf (Step 4)} Finally, to complement this set with a sample of entirely unassociated pairs, we followed the SimLex-999 setup. We paired up the verbs from the 3,153 associated pairs at random. From these random parings, we excluded those that coincidentally occurred elsewhere in USF (and therefore had a degree of association). We sampled the remaining 347 pairs from this resulting set of unassociated pairs.

\noindent {\bf (Output)} The final SimVerb-3500 data set contains 3,500 verb pairs in total, covering all associated verb pairs from USF, and (almost) all top-level VerbNet classes. All pairs were manually checked post-hoc by the authors plus 2 additional native English speakers to verify that the final data set does not contain unknown or invalid verb types.

\paragraph{Frequency Statistics} The 3,500 pairs consist of 827 distinct verbs. 29 top-level VN classes are represented by 3 member verbs, while the three most represented classes cover 79, 85, and 93 member verbs. 40 verbs are not members of any VN class.

We performed an initial frequency analysis of SimVerb-3500 relying on the BNC counts available online \cite{Kilgarriff:1997ijl}.\footnote{https://www.kilgarriff.co.uk/bnc-readme.html} After ranking all BNC verbs according to their frequency, we divided the list into quartiles: Q1 (most frequent verbs in BNC) - Q4 (least frequent verbs in BNC). Out of the 827 SimVerb-3500 verb types, 677 are contained in Q1, 122 in Q2, 18 in Q3, 4 in Q4 ({\em to enroll, to hitchhike, to implode, to whelp}), while 6 verbs are not covered in the BNC list. 2,818 verb pairs contain Q1 verbs, while there are 43 verb pairs with both verbs not in Q1. Further empirical analyses are provided in \cref{s:evaluation}.\footnote{Annotations such as VerbNet class membership, relations between WordNet synsets of each verb, and frequency statistics are available as supplementary material.} 

\section{Word Pair Scoring}
\label{s:annotation}

We employ the Prolific Academic (PA) crowdsourcing platform,\footnote{https://prolific.ac/  (We chose PA for logistic reasons.)} an online marketplace very similar to Amazon Mechanical Turk and to CrowdFlower. 

\subsection{Survey Structure}

Following the SimLex-999 annotation guidelines, we had each of the 3500 verb pairs rated by at least 10 annotators. To distribute the workload, we divided the 3500 pairs into 70 tranches, with 79 pairs each. Out of the 79 pairs, 50 are unique to one tranche, while 20 manually chosen pairs are in all tranches to ensure consistency. The remaining 9 are duplicate pairs displayed to the same participant multiple times to detect inconsistent annotations.

Participants see 7-8 pairs per page. Pairs are rated on a scale of 0-6 by moving a slider. The first page shows 7 pairs, 5 unique ones and 2 from the consistency set. The following pages are structured the same but display one extra pair from the previous page. Participants are explicitly asked to give these duplicate pairs the same rating. We use them as quality control so that we can identify and exclude participants giving several inconsistent answers.

\paragraph{Checkpoint Questions} The survey contains three control questions in which participants are asked to select the most similar pair out of three choices. For instance, the first checkpoint is: \textit{Which of these pairs of words is the *most* similar? 1. to run / to jog 2. to run / to walk 3. to jog / to sweat.} One checkpoint occurs right after the instructions and the other two later in the survey. The purpose is to check that annotators have understood the guidelines and to have another quality control measure for ensuring that they are paying attention throughout the survey. If just one of the checkpoint questions is answered incorrectly, the survey ends immediately and all scores from the annotator in question are discarded.  

\paragraph{Participants}
843 raters participated in the study, producing over 65,000 ratings. Unlike other crowdsourcing platforms, PA collects and stores detailed demographic information from the participants upfront. This information was used to carefully select the pool of eligible participants. We restricted the pool to native English speakers with a 90\% approval rate (maximum rate on PA), of age 18-50, born and currently residing in the US (45\% out of 843 raters), UK (53\%), or Ireland (2\%). 54\% of the raters were female and 46\% male, with the average age of 30. Participants took 8 minutes on average to complete the survey containing 79 questions.
 
\subsection{Post-Processing}

We excluded ratings of annotators who \textit{(a)} answered one of the checkpoint questions incorrectly  (75\% of exclusions); \textit{(b)} did not give equal ratings to duplicate pairs; \textit{(c)} showed suspicious rating patterns (e.g., randomly alternating between two ratings or using one single rating throughout). 
The final acceptance rate was 84\%. We then calculated the average of all ratings from the accepted raters ( $\geq$ 10 ) for each pair. The score was finally scaled linearly from the 0-6 to the 0-10 interval as in \cite{Hill:2015cl}.

\section{Analysis}
\label{s:analysis}

\paragraph{Inter-Annotator Agreement}

We employ two measures. \textbf{IAA-1 (pairwise)} computes the average pairwise Spearman's $\rho$ correlation between any two raters -- a common choice in previous data collection in distributional semantics \cite{Pado:2007emnlp,Reisinger:2010emnlp,Silberer:2014acl,Hill:2015cl}.

A complementary measure would smooth individual annotator effects. For this aim, our \textbf{IAA-2 (mean)} measure compares the average correlation of a human rater with the average of all the other raters. SimVerb-3500 obtains $\rho$ = 0.84 (IAA-1) and $\rho$ = 0.86 (IAA-2), a very good agreement compared to other benchmarks (see Tab.~\ref{tab:gaps}).

\paragraph{Vector Space Models}

We compare the performance of prominent representation models on SimVerb-3500. We  include: (1) unsupervised models that learn from distributional information in text, including the skip-gram negative-sampling model (\textit{SGNS}) with various contexts (\textit{BOW = bag of words; DEPS = dependency contexts}) as in \newcite{Levy:2014acl}, the symmetric-pattern based vectors by \newcite{Schwartz:2015conll}, and count-based PMI-weighted vectors \cite{Baroni2014acl}; (2) Models that rely on linguistic hand-crafted resources or curated knowledge bases. Here, we use sparse binary vectors built from linguistic resources (\textit{Non-Distributional}, \cite{Faruqui:2015acl}), and vectors fine-tuned to a paraphrase database (\textit{Paragram}, \cite{Wieting:2015tacl}) further refined using linguistic constraints (\textit{Paragram+CF}, \cite{Mrksic:2016naacl}). Descriptions of these models are in the supplementary material. 

\begin{table}
\begin{center}
\def\arraystretch{0.9}
\scriptsize
\begin{tabular}{l cccc}
\toprule
  {\bf Eval set} & {IAA-1} & {IAA-2} & {\sc All} &  {\sc Text}\\ 
  \cmidrule(lr){2-5}
  {\sc WSim} & {0.67} & {0.65} & {0.79} & {0.79} \\
  {(203)} & {} & {} & {SGNS-BOW} & {SGNS-BOW}\\
  {\sc SimLex} & {0.67} & {0.78} & {0.74} & {0.56}\\
   {(999)} & {} & {} & {Paragram+CF} & {SymPat+SGNS}\\
  \midrule
  {\sc SL-222} & {0.72} & {-} & {0.73} & {0.58}\\
  {(222)} & {} & {} & {Paragram+CF} & {SymPat}\\
  \midrule
  {\sc SimVerb} & {0.84} & {0.86} & {0.63} & {0.36} \\
  {(3500)} & {} & {} & {Paragram+CF} & {SGNS-DEPS}\\
\bottomrule
\end{tabular}
\end{center}
\caption{An overview of word similarity evaluation benchmarks. {\sc All} is the current best reported score on each data set across all models (including the models that exploit curated knowledge bases and hand-crafted lexical resources, see supplementary material). {\sc Text} denotes the best reported score for a model that learns solely on the basis of distributional information. All scores are Spearman's $\rho$ correlations.}
\label{tab:gaps}
\vspace{-1.0em}
\end{table}

\begin{table*}
\begin{footnotesize}
\def\arraystretch{0.9}
\begin{center}
\hspace*{-0.5cm}\begin{tabular}{l c | cc |cc}
\toprule
  {\bf Model} & {\sc SV-3500} & {\sc CV-222}  & {\sc SL-222}  & {\sc DEV-500} & {\sc TEST-3000}  \\ 
  \cmidrule(lr){2-6}
  {\sc SGNS-BOW-PW} (d=300) &  {0.274} & {0.279} & {0.328} & {0.333} & {0.265} \\
  {\sc SGNS-DEPS-PW} (d=300) & {0.313} & {0.314} & {0.390} & {0.401} & {0.304}\\
  {\sc SGNS-UDEPS-PW} (d=300) & {0.259} & {0.262} & {0.347} & {0.313} & {0.250} \\
  {\sc SGNS-BOW-8B} (d=500) & {0.348} & {0.343} & {0.307} & {0.378} & {0.350} \\
  {\sc SGNS-DEPS-8B} (d=500) & {0.356} & {0.347} & {0.385} & {0.389} & {0.351} \\
  \midrule
  {\sc SymPat-8B} (d=500) & {0.328} & {0.336} & {0.544} & {0.276} & {0.347} \\
  \midrule
  {\sc Count-SVD} (d=500) & {0.196} & {0.200} & {0.059} & {0.259} & {0.186} \\
  \midrule
  \midrule
  {\sc Non-distributional } & {0.596} & {0.596} & {0.689}  & {0.632} & {0.600}\\
  {\sc Paragram} (d=25) & {0.418} & {0.432} & {0.531} & {0.443} & {0.433} \\
  {\sc Paragram} (d=300) & {0.540} & {0.528} & {0.590} & {0.525} & {0.537}  \\
  {\sc Paragram}+CF (d=300) & {0.628} & {0.625} & {0.727} & {0.611} & {0.624}\\
\bottomrule
\end{tabular}
\end{center}
\end{footnotesize}
\caption{Evaluation of state-of-the representation learning models on the full SimVerb-3500 set (SV-3500), the Simlex-999 verb subset containing 222 pairs (SL-222), cross-validated subsets of 222 pairs from SV-3500 (CV-222), and the SimVerb-3500 development (DEV-500) and test set (TEST-3000).}
\label{tab:slsv}
\label{tab:devc}
\vspace{-1em}
\end{table*}

\paragraph{Comparison to SimLex-999 (SL-222)}

170 pairs from SL-222 also appear in SimVerb-3500. The correlation between the two data sets calculated on the shared pairs is $\rho=0.91$. This proves, as expected, that the ratings are consistent across the two data sets.

Tab.~\ref{tab:slsv} shows a comparison of models' performance on SimVerb-3500 against SL-222. Since the number of evaluation pairs may influence the results, we ideally want to compare sets of equal size for a fair comparison. Picking one random subset of 222 pairs would bias the results towards the selected pairs, and even using 10-fold cross-validation we found variations up to 0.05 depending on which subsets were used. Therefore, we employ a 2-level 10-fold cross-validation where new random subsets are picked in each iteration of each model. The numbers reported as \textit{CV-222} are averages of these ten 10-fold cross-validation runs. The reported results come very close to the correlation on the full data set for all models.

Most models perform much better on SL-222, especially those employing additional databases or linguistic resources. The performance of the best scoring Paragram+CF model is even on par with the IAA-1 of 0.72. The same model obtains the highest score on SV-3500 ($\rho=0.628$), with a clear gap to IAA-1 of 0.84.  We attribute these differences in performance largely to SimVerb-3500 being a more extensive and diverse resource in terms of verb pairs. 

\paragraph{Development Set}

A common problem in scored word pair datasets is the lack of a standard split to development and test sets. Previous works often optimise models on the entire dataset, which leads to overfitting \cite{Faruqui:2016arxiv} or use custom splits, e.g., 10-fold cross-validation \cite{Schwartz:2015conll}, which make results incomparable with others. The lack of standard splits stems mostly from small size and poor coverage -- issues which we have solved with SimVerb-3500. 

Our development set contains 500 pairs, selected to ensure a broad coverage in terms of similarity ranges (i.e., non-similar and highly similar pairs, as well as pairs of medium similarity are represented) and top-level VN classes (each class is represented by at least 1 member verb). The test set includes the remaining 3,000 verb pairs. The performances of representation learning architectures on the dev and test sets are reported in Tab.~\ref{tab:devc}. The ranking of models is identical on the test and the full SV-3500 set, with slight differences in ranking on the development set.

\section{Evaluating Subsets}
\label{s:evaluation}

The large coverage and scale of SimVerb-3500 enables model evaluation based on selected criteria.  In this section, we showcase a few example analyses.

\begin{figure}[t]
\begin{center}
\includegraphics[width=0.96\linewidth]{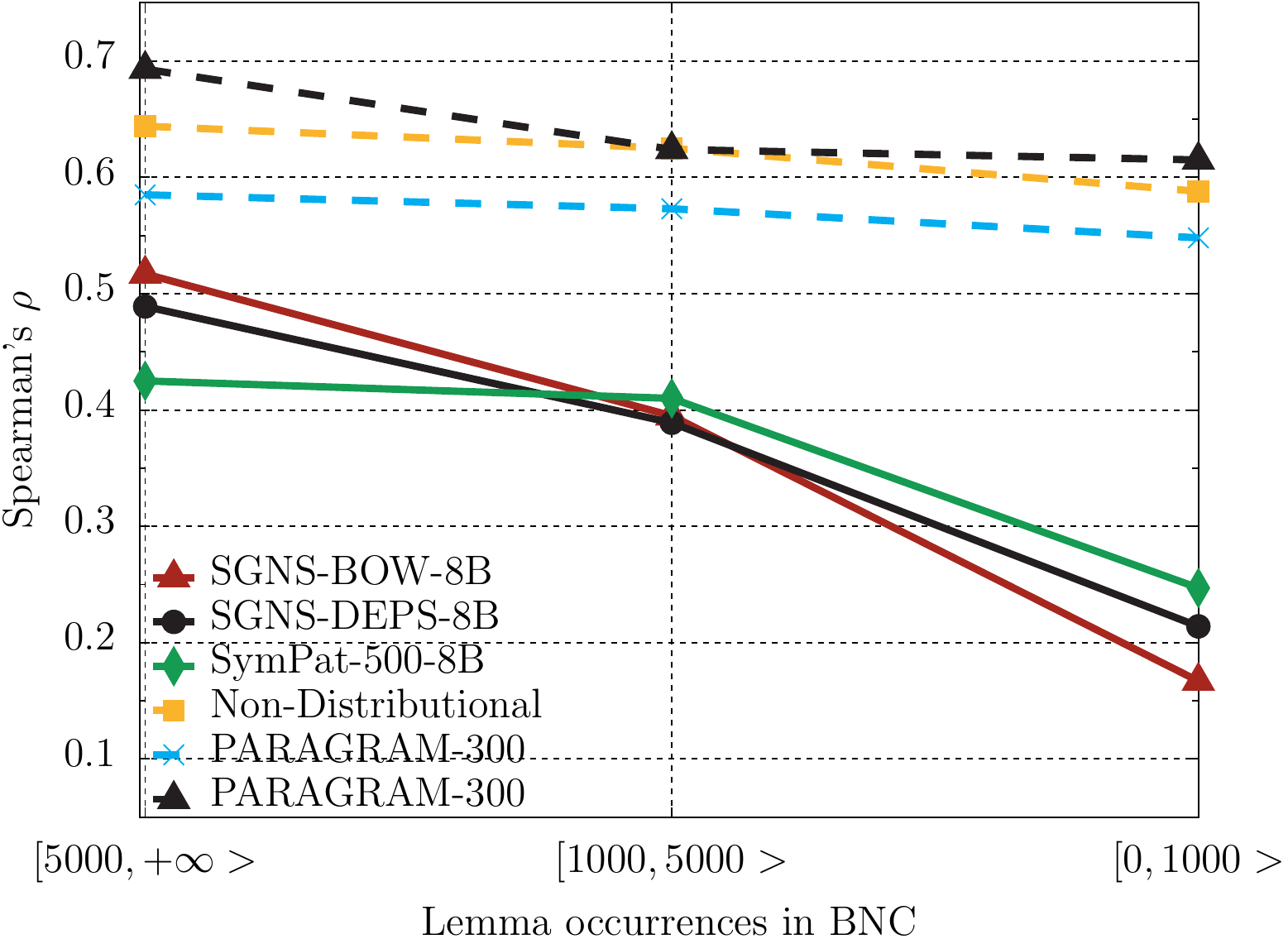}
\vspace{-0.1cm}
\caption{Subset-based evaluation, where subsets are created based on the frequency of verb lemmas in the BNC corpus. Each of the three frequency groups contains 390-490 verb pairs. To be included in each group it is required that both verbs in a pair are contained in the same frequency interval (x axis).}
\vspace{-0.6cm}
\label{fig:bncfreq}
\end{center}
\end{figure}

\paragraph{Frequency} In the first analysis, we select pairs based on their lemma frequency in the BNC corpus and form three groups, with 390-490 pairs in each group (Fig.~\ref{fig:bncfreq}). The results from Fig.~\ref{fig:bncfreq} suggest that the performance of all models improves as the frequency of the verbs in the pair increases, with much steeper curves for the purely distributional models (e.g., SGNS and SymPat). The non-distributional non data-driven model of \newcite{Faruqui:2015acl} is only slightly affected by frequency.  

\begin{figure}[t]
\begin{center}
\includegraphics[width=0.96\linewidth]{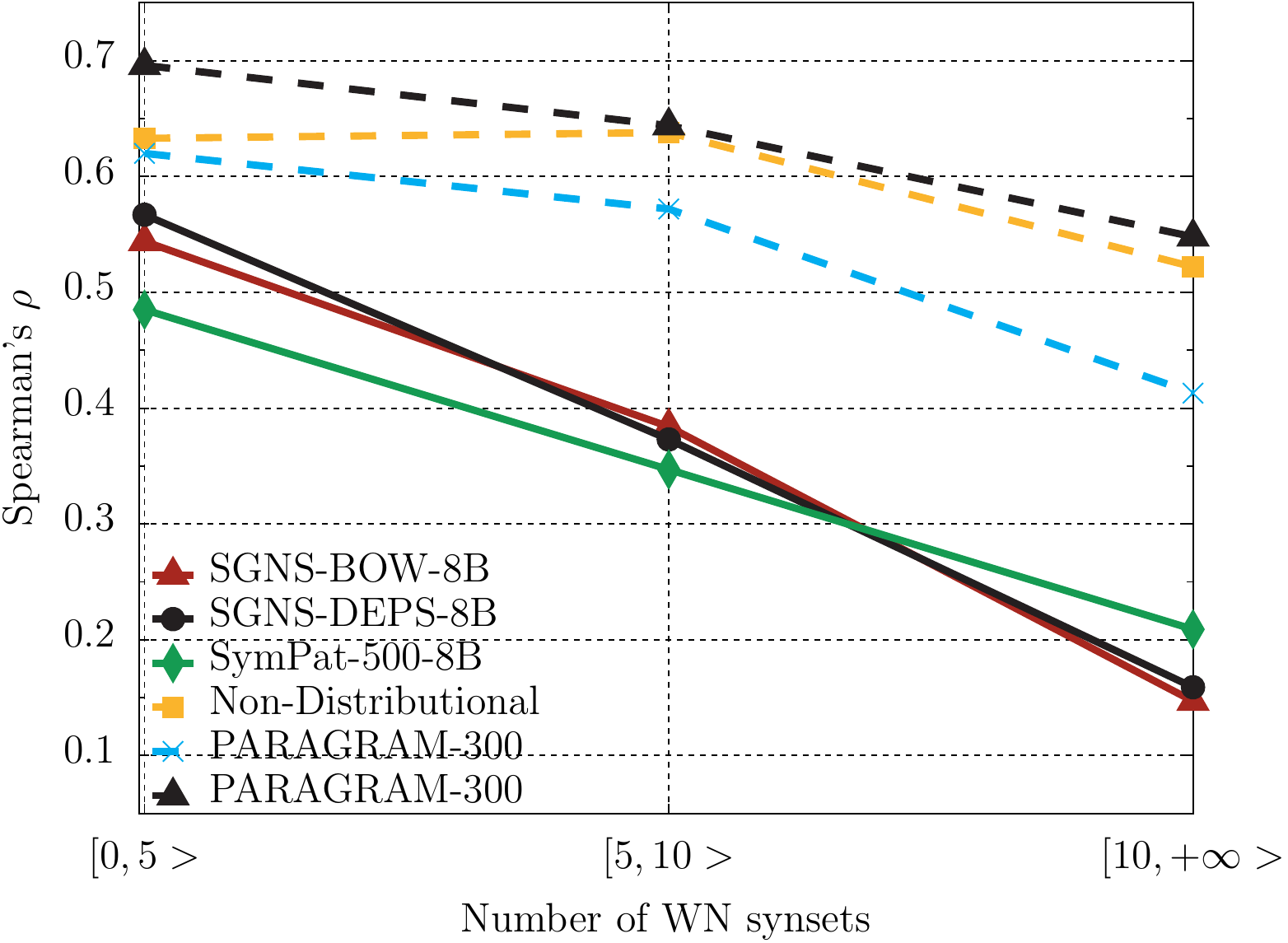}
\vspace{-0.1cm}
\caption{Subset-based evaluation, where subsets are created based on the number of synsets in WordNet (x axis). To be included in each subset it is required that both verbs in a pair have the number of synsets in the same interval.}
\vspace{-0.5cm}
\label{fig:synsets}
\end{center}
\end{figure}

\paragraph{WordNet Synsets} Intuitively, representations for verbs with more diverse usage patterns are more difficult to learn with statistical models. To examine this hypothesis, we resort to WordNet \cite{Miller:1995cacm}, where different semantic usages of words are listed as so-called \textit{synsets}. Fig.~\ref{fig:synsets} shows a clear downward trend for all models, confirming that polysemous verbs are more difficult for current verb representation models. Nevertheless, approaches which use additional information beyond corpus co-occurrence are again more robust. Their performance only drops substantially for verbs with more than 10 synsets, while the performance of other models deteriorates already when tackling verbs with more than 5 synsets.

\begin{table}[t]
\begin{center}
\scriptsize
\hspace*{-0.25cm}\begin{tabular}{l ccccc}
\toprule
  {\bf Model} & {\sc \#13} & {\sc \#31} & {\sc \#37} & {\sc \#45} &  {\sc \#51}  \\ 
  \cmidrule(lr){2-6}
  {\sc SGNS-BOW-8B} & {0.210} & {0.308} & {0.352} & {0.270} & {0.170}\\
  {\sc SGNS-DEPS-8B} & {0.289} & {0.270} & {0.306} & {0.238} & {0.225}\\
  \midrule
  {\sc SymPat-8B} (d=500) & {0.171} & {0.320} & {0.143} & {0.195} & {0.113} \\
  \midrule
  \midrule
  {\sc Non-distr}   & {0.571} & {0.483} & {0.372} & {0.501} & {0.499} \\
  {\sc Paragram} (d=300) & {0.571} & {0.504} & {0.567} & {0.531} & {0.387}\\
  {\sc Paragram+CF}& {0.735} & {0.575} & {0.666} & {0.622} & {0.614}\\
\bottomrule
\end{tabular}
\caption{Spearman's $\rho$ correlation between human judgments and model's cosine similarity by VerbNet Class. We chose classes \#13 \textit{Verbs of Change of Possession}, \#31 \textit{Verbs of Psychological State}, \#37 \textit{Verbs of Communication}, \#45 \textit{Verbs of Change of State}, and \#51 \textit{Verbs of Motion} as examples. All are large classes with more than 100 pairs each, and the frequencies of member verbs are distributed in a similar way.}
\label{tab:verbnet}
\end{center}
\vspace{-0.3em}
\end{table}

\begin{table}[t]
\begin{center}
\def\arraystretch{1.0}
\scriptsize
\begin{tabular}{l ccc}
\toprule
  {\bf Model}  & {\sc NR} & {\sc SYN} & {\sc HYP} \\ 
  \cmidrule(lr){2-4}
  {\sc SGNS-BOW-PW} (d=300) & {0.096} & {0.288} & {0.292} \\
  {\sc SGNS-DEPS-PW} (d=300) & {0.132} & {0.290} & {0.336} \\
  {\sc SGNS-BOW-8B} (d=500) & {0.292} & {0.273} & {0.338} \\
  {\sc SGNS-DEPS-8B } (d=500) & {0.157} & {0.323} & {0.378}\\
  \midrule  
  {\sc SymPat-8B-dense} (d=300) & {0.225} & {0.182} & {0.265} \\
  {\sc SymPat-8B-dense} (d=500) & {0.248} & {0.260} & {0.251}\\
  \midrule
  {\sc Non-distributional }  & {0.126} & {0.379} & {0.488}\\
  {\sc Paragram} (d=300) & {0.254} & {0.356} & {0.439} \\
  {\sc Paragram}+CF (d=300) & {0.250} & {0.417} & {0.475} \\
\bottomrule
\end{tabular}
\caption{Spearman's $\rho$ correlation between human judgments and model's cosine similarity based on pair relation type. Relations are based on WordNet, and included in the dataset. The classes are of different size, 373 pairs with no relation (\textit{NR}), 306 synonym (\textit{SYN}) pairs, and 800 hyper/hyponym (\textit{HYP}) pairs. Frequencies of member verbs are distributed in a similar way.}
\label{table:relations}
\end{center}
\vspace{-1.3em}
\end{table}

\paragraph{VerbNet Classes} Another analysis enabled by SimVerb-3500 is investigating the connection between VerbNet classes and human similarity judgments. We find that verbs in the same top-level VerbNet class are often not assigned high similarity score. Out of 1378 pairs where verbs share the top-level VerbNet class, 603 have a score lower than 5.
Tab.~\ref{tab:verbnet} reports scores per VerbNet class. When a verb belongs to multiple classes, we count it for each class (see Footnote~2). We run the analysis on the five largest VN classes, each with more than 100 pairs with paired verbs belonging to the same class. 

The results indicate clear differences between classes (e.g., Class 31 vs Class 51), and suggest that further developments in verb representation learning should also focus on constructing specialised representations at the finer-grained level of VN classes.

\paragraph{Lexical Relations} SimVerb-3500 contains relation annotations (e.g., \textit{antonyms}, \textit{synonyms}, \textit{hyper-/hyponyms}, \textit{no relation}) for all pairs extracted automatically from WordNet. Evaluating per-relation subsets, we observe that some models draw their strength from good performance across different relations. Others have low performance on these pairs, but do very well on synonyms and hyper-/hyponyms. Selected results of this analysis are in Tab.~\ref{table:relations}.\footnote{
Evaluation based on Spearman's $\rho$ may be problematic with certain categories, e.g., with antonyms. It evaluates pairs according to their ranking; for antonyms the ranking is arbitrary - every antonym pair should have a very low rating, hence they are not included in Tab.~\ref{table:relations}. A similar effect occurs with highly ranked synonyms, but to a much lesser degree than with antonyms.}

\paragraph{Human Agreement} 
Motivated by the varying performance of computational models regarding frequency and ambiguous words with many synsets, we analyse what disagreement effects may be captured in human ratings. We therefore compute the average standard deviation of ratings per subset: $ avgstdd(S) = \frac{1}{n} \sum_{p \in S} \sigma (r_p)$, where \textit{S} is one subset of pairs, \textit{n} is the number of pairs in this subset, \textit{p} is one pair, and \textit{$r_p$} are all human ratings for this pair.

While the standard deviation of ratings is diverse for individual pairs, overall the average standard deviations per subset are almost identical. For both the frequency and the WordNet synset analyses it is around {$\approx$1.3} across all subsets, and with only little difference for the subsets based on VerbNet. The only subsets where we found significant variations is the grouping by relations, where ratings tend to be more similar especially on antonyms (\textit{0.86}) and pairs with no relation (\textit{0.92}), much less similar on synonyms (\textit{1.34}) and all other relations (\textit{$\approx$1.4}). These findings suggest that humans are much less influenced by frequency or polysemy in their understanding of verb semantics compared to computational models.

\section{Conclusions}
\label{s:conclusion}

SimVerb-3500 is a verb similarity resource for analysis and evaluation that will be of use to researchers involved in understanding how humans or machines represent the meaning of verbs, and, by extension, scenes, events and full sentences. The size and coverage of syntactico-semantic phenomena in SimVerb-3500 makes it possible to compare the strengths and weaknesses of various representation models via statistically robust analyses on specific word classes. 

To demonstrate the utility of SimVerb-3500, we conducted a selection of analyses with existing representation-learning models. One clear conclusion is that distributional models trained on raw text (e.g. SGNS) perform very poorly on low frequency and highly polysemous verbs. This degradation in performance can be partially mitigated by focusing models on more principled distributional contexts, such as those defined by symmetric patterns. More generally, the finding suggests that, in order to model the diverse spectrum of verb semantics, we may require algorithms that are better suited to fast learning from few examples \cite{lake2011one}, and have some flexibility with respect to sense-level distinctions \cite{reisinger2010multi,vilnis2014word}. In future work we aim to apply such methods to the task of verb acquisition. 

Beyond the preliminary conclusions from these initial analyses, the benefit of SimLex-3500 will become clear as researchers use it to probe the relationship between architectures, algorithms and representation quality for a wide range of verb classes. Better understanding of how to represent the full diversity of verbs should in turn yield improved methods for encoding and interpreting the facts, propositions, relations and events that constitute much of the important information in language. 

\section*{Acknowledgments} This work is supported by the ERC Consolidator Grant LEXICAL (648909).

\bibliographystyle{emnlp2016}
\bibliography{main.bib} 

\newpage

\twocolumn[\begin{center} 
        \Large  SimVerb-3500: A Large-Scale Evaluation Set of Verb Similarity \\ Supplementary Material
\end{center}]
\section*{Vector Space Models}
\label{a:vsm}
\subsection*{Unsupervised Text-Based Models}
\label{ss:unsupervised}
These models mainly learn from co-occurrence statistics in large corpora, therefore to facilitate the generality of our results, we evaluate them on two different corpora. With \textbf{8B} we refer to the corpus produced by the \texttt{word2vec} script, consisting of 8 billion tokens from various sources \cite{Mikolov:2013iclr}.\footnote{https://code.google.com/archive/p/word2vec/}  With \textbf{PW} we refer to the English Polyglot Wikipedia corpus \cite{AlRfou:2013conll}.\footnote{https://sites.google.com/site/rmyeid/projects/polyglot} $d$ denotes the embedding dimensionality, and $ws$ is the window size in case of bag-of-word contexts.
The models we consider are as follows:
\paragraph{SGNS-BOW (PW, 8B)} Skip-gram with negative sampling (SGNS) \cite{Mikolov:2013iclr,Mikolov:2013nips} trained with bag-of-words (BOW) contexts; $d=500$, $ws=2$ on 8B as in prior work \cite{Melamud:2016naacl,Schwartz:2016naacl}. $d=300$, $ws=2$ on PW as in prior work \cite{Levy:2014acl,Vulic:2016acluniversal}. 
\paragraph{SGNS-UDEP (PW)} SGNS trained with universal dependency\footnote{http://universaldependencies.org/ (version 1.2)} (UD) contexts following the setup of \cite{Levy:2014acl,Vulic:2016acluniversal}. The PW data were POS-tagged with universal POS (UPOS) tags \cite{Petrov:2012lrec} using  TurboTagger \cite{Martins:2013acl}\footnote{http://www.cs.cmu.edu/~ark/TurboParser/}, trained using default settings without any further parameter fine-tuning (SVM MIRA with 20 iterations) on the {\sc train+dev} portion of the UD treebank annotated with UPOS tags. The data were then parsed using the graph-based Mate parser v3.61 \cite{Bohnet:2010coling}.\footnote{https://code.google.com/archive/p/mate-tools/} $d=300$ as in \cite{Vulic:2016acluniversal}
\paragraph{SGNS-DEP (8B)} Another variant of a dependency-based SGNS model is taken from the recent work of \newcite{Schwartz:2016naacl}, based on \newcite{Levy:2014acl}. The 8B corpus is parsed with labeled Stanford dependencies \cite{Marneffe:2008sd}, the Stanford POS Tagger \cite{Toutanova:2003naacl} and the stack version of the MALT parser \cite{Goldberg:2012coling} are used; $d=500$ as in prior work \cite{Schwartz:2016naacl}.

All other parameters of all SGNS models are set to the standard settings: the models are trained with stochastic gradient descent, global learning rate of 0.025, subsampling rate $1e-4$, $15$ epochs.

\paragraph{SymPat (8B)} A template-based approach to vector space modeling introduced by \newcite{Schwartz:2015conll}. Vectors are trained based on co-occurrence of words in symmetric patterns \cite{Davidov:2006acl}, and an antonym detection mechanism is plugged in the representations. We use pre-trained dense vectors ($d=300$ and $d=500$) with the antonym detector enabled, available online.\footnote{http://www.cs.huji.ac.il/\textasciitilde roys02/papers/sp\_embeddings/}

\paragraph{Count-SVD} Traditional count-based vectors using PMI weighting and SVD dimensionality reduction (\textit{ws = 2; d = 500}). This is the best performing reduced count-based model from \newcite{Baroni2014acl}, vectors were obtained online.\footnote{http://clic.cimec.unitn.it/composes/semantic-vectors.html}

\subsection*{Models Relying on External Resources}
\label{ss:external}
\paragraph{Non-Distributional} Sparse binary vectors built from a wide variety of hand-crafted linguistic resources, e.g., WordNet, Supersenses, FrameNet, Emotion and Sentiment lexicons, Connotation lexicon, among others \cite{Faruqui:2015acl}.\footnote{https://github.com/mfaruqui/non-distributional}
\paragraph{Paragram} \newcite{Wieting:2015tacl} use the Paraphrase Database (PPDB) \cite{Ganitkevitch2013naacl} word pairs to learn word vectors which emphasise paraphrasability. They do this by fine-tuning, also known as retro-fitting \cite{Faruqui:2015naacl}, \texttt{word2vec} vectors using a SGNS inspired objective function designed to incorporate the PPDB semantic similarity constraints. Two variants are available online: $d=25$ and $d=300$.\footnote{http://ttic.uchicago.edu/$\sim$wieting/}
\paragraph{Paragram+CF} \newcite{Mrksic:2016naacl} suggest another variant of the retro-fitting procedure called counter-fitting (CF) which further improves the Paragram vectors by injecting antonymy constraints from PPDB v2.0 \cite{Pavlick:2015acl} into the final vector space. $d=300$.\footnote{https://github.com/nmrksic/counter-fitting}

\end{document}